\documentclass{article}
\usepackage{spconf,amsmath,graphicx}
\usepackage{amsfonts}
\usepackage{makecell}
\usepackage{booktabs}

\usepackage{algorithm}
\usepackage{algpseudocode}
\usepackage{multicol}
\usepackage{multirow}
\usepackage{rotating}
\usepackage{color}

\usepackage{enumitem}
\setlist{nosep, leftmargin=14pt}

\usepackage{mwe} 

\title{UNetVL: Enhancing 3D Medical Image Segmentation with Chebyshev KAN Powered Vision-LSTM}
\name{Xuhui Guo$^{1}$, Tanmoy Dam$^{2}$, Rohan Dhamdhere$^{2}$, Gourav Modanwal$^{2}$, Anant Madabhushi$^{2}$}
\address{\small
    $^1$Electrical Engineering and Computer Science Department, University of Michigan, Ann Arbor, USA\\
    \small
    $^2$Wallace H. Coulter Department of Biomedical Engineering, Georgia Institute of Technology and Emory University, Atlanta, USA\\[2ex]}
\begin{document}
%
\maketitle
\begin{abstract}

3D medical image segmentation has progressed considerably due to Convolutional Neural Networks (CNNs) and Vision Transformers (ViTs), yet these methods struggle to balance long-range dependency acquisition with computational efficiency. To address this challenge, we propose UNETVL (\textbf{U}-\textbf{Net} \textbf{V}ision-\textbf{L}STM), a novel architecture that leverages recent advancements in temporal information processing. UNETVL incorporates Vision-LSTM (ViL) for improved scalability and memory functions, alongside an efficient Chebyshev Kolmogorov-Arnold Networks (KAN) to handle complex and long-range dependency patterns more effectively. We validated our method on the ACDC and AMOS2022 (post challenge Task 2) benchmark datasets, showing a significant improvement in mean Dice score compared to recent state-of-the-art approaches, especially over its predecessor, UNETR, with increases of 7.3\% on ACDC and 15.6\% on AMOS, respectively. Extensive ablation studies were conducted to demonstrate the impact of each component in UNETVL, providing a comprehensive understanding of its architecture. Our code is available at https://github.com/tgrex6/UNETVL, facilitating further research and applications in this domain.

\end{abstract}

\section{Introduction}
\label{sec:intro}

The incorporation of artificial intelligence (AI) into medical image process has profoundly revolutionized patient care, providing a more efficient and accessible approach to improving diagnostic precision and prognostic outcomes. Accurate segmentation, in particular, acts as a fundamental step in biomedical image analysis, delineating key anatomical structures or regions of interest to facilitate interpretation by clinicians and researchers \cite{hatamizadeh2022unetr}. Researchers have thoroughly investigated numerous deep learning-based segmentation techniques following the emergence of a new age in AI, attributed to advancements in graphical processing units (GPUs). In the past decade, Convolutional Neural Networks (CNNs), particularly U-shaped encoder-decoder architectures \cite{ronneberger2015u}, have been fundamental to the most advanced techniques in this field. Nonetheless, although CNNs are proficient at extracting local characteristics, they have difficulties in capturing long-range dependencies due to the static nature of the receptive fields in convolutional layers \cite{hatamizadeh2022unetr}. 

In recent years, Vision Transformers (ViTs) have facilitated dynamic learning of relationships between distant regions in input data by employing self-attention mechanisms, thereby being able to address the challenge of modeling spatial dependencies. A notable example is the UNETR model \cite{hatamizadeh2022unetr}, which leverages this advantage by incorporating ViT layers as the feature extractor for the encoder. This approach has demonstrated superior efficacy across several 3D medical image segmentation tasks compared to CNN-based methods. However, ViTs face significant computational challenges, particularly in high-resolution tasks, due to their quadratic complexity relative to input size. This limitation poses a considerable barrier in medical imaging applications, where high-resolution three-dimensional volumes are typically the standard. The computational burden of ViTs becomes pronounced especially when processing such large, detailed medical images.

In the latest development, the original team behind LSTM introduced xLSTM \cite{beck2024xlstm}, a significant improvement that incorporates new gating mechanisms and memory structures to enhance both performance and scalability when dealing with large-scale data. Vision-LSTM (ViL) \cite{alkin2024vision}, built on xLSTM, consisting of stacked mLSTM blocks, which follow the bidirectional processing mechanism, is able to efficiently capture both contextual and spatial information, making it highly suited for complex image segmentation tasks, where preserving spatial relationships is essential.

In this work, we propose to replace the ViT layers in the UNETR model with ViL layers, introducing what we call UNET ViL (UNETVL) architecture. In addition, we modify the up and down projection layer in ViL and leverage the stronger nonlinear representation capabilities of the Kolmogorov–Arnold Network (KAN) \cite{liu2024kan} compared to Multi-Layer Perceptrons (MLP) to help capture subtle details in medical images. We validate the effectiveness of our method on 2 multi-structure and multi-organ 3D CT and MRI segmentation benchmarks, namely Automated Cardiac Diagnosis Challenge (ACDC) \cite{bernard2018deep} and AMOS2022 (post challenge Task 2) \cite{NEURIPS2022_ee604e1b} as suggested in the nnU-Net revisited work \cite{isensee2024nnu}. The key contributions of this work are as follows: 
\begin{itemize}
    \item We investigate the efficacy of ViL as an encoder in a 3D medical image segmentation framework based on UNETR \cite{hatamizadeh2022unetr}, making the model more efficient in memory without sacrificing segmentation accuracy.
    \item In addition, we have incorporated Chebyshev KAN \cite{ss2024chebyshev} to replace the MLP-based univariate function in our model. This modification aims to capture more complex encoding distributions in segmentation tasks.
    \item We validate our proposed method, UNETVL, using two benchmark datasets in both CT and MRI imaging domain to evaluate its performance against current state-of-the-art (SOTA) methods. Additionally, we have conducted ablation studies to analyze the impact of each component on the overall performance of our model.

    
\end{itemize}

\begin{figure*}[!ht]
    \centering
    \includegraphics[width=\textwidth]{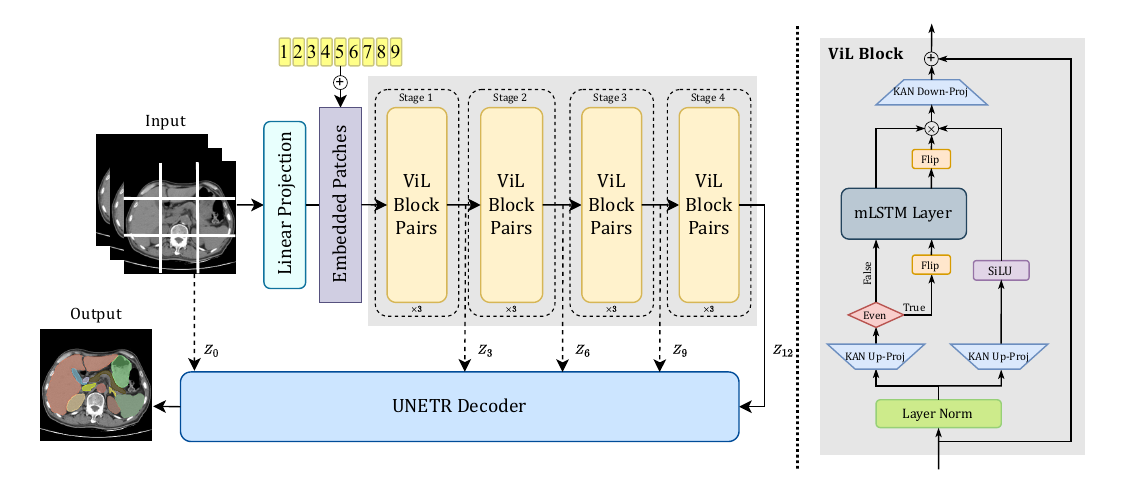}
    \caption{Left: Overall architecture of UNETVL; Right: The internal structure of the ViL blocks. }
    \label{model}
\end{figure*}

\section{Methodology}
\label{sec:format}

\subsection{Architecture Overview}
We present the overall architecture of our UNETVL model in Fig.~\ref{model}. Building upon the UNETR structure~\cite{hatamizadeh2022unetr}, UNETVL employs an encoder-decoder network that efficiently captures both local features and long-range contexts. The model processes a 3D oversampled instance with resolution $(H, W, D)$ and $C$ input channels, applying preprocessing and data augmentation steps from the nnUNet framework\cite{isensee2021nnu}. The input is divided into non-overlapping patches of shape $(P, P, P)$, which are linearly projected into $N = (\frac{H}{P} \times \frac{W}{P} \times \frac{D}{P})$ patch tokens. Learnable positional embeddings are added to maintain spatial information. These tokens are sequentially processed in a $K$-dimensional embedding space through multiple Vision-LSTM (ViL) block pairs, generating intermediate feature representations at various scales (e.g., $z_3, z_6, z_9, z_{12}$), each with shape $(N, K)$. The embedding dimension $K$ remains constant throughout the network depth. Finally, these extracted latent representations are passed to a CNN-based decoder via skip connections, which merges the multi-scale features to produce the final segmentation output. This architecture combines the strengths of Vision Transformers for capturing global context with the efficiency of CNNs for local feature processing, resulting in a powerful model for 3D medical image segmentation tasks.

 
\subsection{ViL Block}
\label{ssec:subhead}
The ViL blocks in our model comprise alternating mLSTM layers that process patch tokens bidirectionally: forward (from top-left to bottom-right) and backward (from bottom-right to top-left). At the core of these blocks lies the mLSTM architecture, which enhances memory capabilities and spatial information retrieval by replacing the scalar memory cell with a $d \times d$ matrix. This modification allows the model to capture more complex data relationships and patterns within a single time step. The mLSTM design draws inspiration from the Query-Key-Value (QKV) attention mechanism in transformers but diverges in its computational approach. This adaptation enables parallelization of LSTM operations, significantly improving computational efficiency and allowing the model to scale more effectively to large datasets. 

\subsection{Chebyshev KAN}
\label{ssec:subhead}
As a promising alternative to MLP, Kolmogorov-Arnold Network (KAN) \cite{liu2024kan} represents a variant of the traditional feedforward neural network with a distinctive twist: its activation functions are learnable and are moved from the nodes to the edges. Chebyshev KAN \cite{ss2024chebyshev} is an innovative framework that combines KAN with the efficient and intuitive Chebyshev polynomials to enhance the performance of the original B-spline approach. The core of this framework are Chebyshev polynomials, a class of orthogonal polynomials defined on the interval [-1, 1], which are well-suited for function approximation. The Chebyshev KAN layer, built on these polynomials, offers a novel alternative to original B-splines, addressing their limitations in both performance and ease of use. Integrating Chebyshev KAN into existing projects is straightforward, requiring only a single line of code to implement. Our modified ViL block incorporates Chebyshev KAN transformations to help improve the flexibility and adaptability of the architecture to approximate high-dimensional feature representation capabilities and we have conducted ablation experiment to prove the feasibility.

For implementation in our up and down projection layer, we just pass the feature embedding of shape \((N, K)\) through the Chebyshev KAN layer, where the Chebyshev polynomials \( \mathbf{T} \in \mathbb{R}^{N \times K \times (\text{degree}+1)} \) is computed recursively as \( T_m(x) \) are  defined as \( T_0(x) = 1 \), \( T_1(x) = x \), and \( T_m(x) = 2xT_{m-1}(x) - T_{m-2}(x) \) for \( m \geq 2 \). A tensor of learnable Chebyshev coefficients \( \mathbf{C} \in \mathbb{R}^{K \times \text{output\_dim} \times (\text{degree}+1)} \) is introduced, serving as trainable parameters. The layer's output \( \mathbf{y} \in \mathbb{R}^{N \times \text{output\_dim}} \) is computed using Einstein summation as 

\[
y_{no} = \sum_{i=1}^{K} \sum_{j=0}^{\text{degree}} T_{nij} \cdot C_{ioj}
\]

\noindent where \( n \), \( i \), \( o \), and \( j \) index the patch, input dimensions, output dimensions, and polynomial degree, respectively.

\section{Experiments}
\label{sec:pagestyle}
\textbf{Datasets.} We evaluate the effectiveness of our approach on two public 3D segmentation benchmarks, both involving multiple structures or multiple organs: the Automated Cardiac Diagnosis Challenge (ACDC) \cite{bernard2018deep} and the AMOS2022 post-challenge Task 2 \cite{NEURIPS2022_ee604e1b}. These benchmarks were selected based on recommendations from the nnU-Net revisited study \cite{isensee2024nnu} for low intra-method segmentation result standard deviation while effectively differentiating between methods, as indicated by a high inter-method standard deviation. We employ 200 cardiac MRI from ACDC with 3 anatomies manually annotated and 360 multi-contrast abdominal CT from AMOS 2022 post-challenge Task 2 with 15 anatomies manually annotated for abdominal multi-organ segmentation. For ACDC dataset, the oversampled instance size is [16, 256, 224]. For AMOS 2022 post-challenge Task 2, the oversampled input image volume size is set as [128, 128, 128].

\noindent\textbf{Implementation Details.} We implemented UNETVL within the widely-used nnU-Net framework \cite{isensee2021nnu}. Primarily, nnU-Net’s modular design allows us to focus on implementing the new network architecture while keeping other aspects, such as image preprocessing and data augmentation, consistent. Following the call for rigorous validation, such a setup ensures a fair comparison of UNETVL with other methods under uniform conditions, with the network architecture as the sole variable. All experiments were conducted on an Amazon EC2 G5 instance with 4 NVIDIA A10G Tensor Core GPUs. All models were trained with a batch size of 4, utilizing the AdamW optimizer with initial learning rate of \(1e^{-4}\) and a PolynomialLR scheduler to adjust learning rate during training for 200 and 600 epochs respectively. We employed the Dice similarity coefficient as the evaluation metric to quantify the overlap between the predicted segmentation and the ground truth labels. We report the average results from 5-fold cross-validation with a rotating validation set.

\begin{figure*}[!ht]
    \centering
    \includegraphics[width=0.94\textwidth]{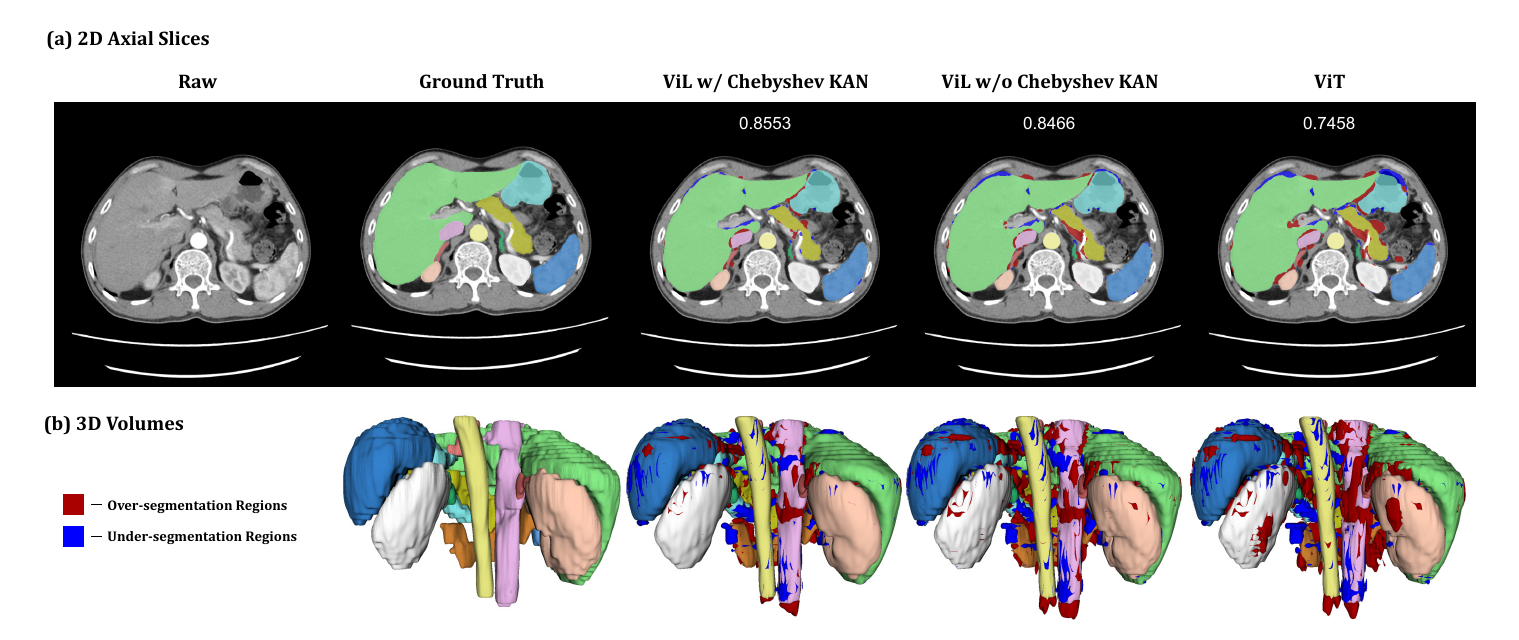}
    \caption{Qualitative comparison of multi-organ segmentation results for models based on UNETR architecture on AMOS 2022 post-challenge Task 2 dataset. The ViL with Chebyshev KAN method shows the best segmentation quality compared to the ground-truth.}
    \label{result}
\end{figure*}

\section{Results}
\label{sec:typestyle}

\textbf{Evaluation on ACDC and AMOS.} As shown in Table \ref{q1}, when comparing the performance of the ViL architecture with Chebyshev KAN to the UNETR baseline, a significant improvement is observed. On the ACDC dataset, the Dice score improves by approximately 7.3\% (from 85.34\% in UNETR to 91.59\% in UNETVL with KAN). Similarly, for the AMOS dataset, the Dice score increases by about 15.6\% (from 76.59\% in UNETR to 88.57\% in UNETVL with KAN). Though UNETVL's results are not the highest when compared with other listed SOTA models on ACDC dataset, its strength lies in addressing over-segmentation issue, which is common in complex anatomical areas. The integration of Chebyshev KAN allows ViL to capture finer details, leading to more precise boundaries and cleaner segmentation. Notably, this approach achieves the highest mean Dice score on the AMOS 2022 post-challenge Task 2 dataset, demonstrating its capability in handling complex medical images with real-world clinical settings.

\begin{table}[!h]
\centering
\tiny
\resizebox{\columnwidth}{!}{ 
\begin{tabular}{l c c c|c}
\toprule
& \textbf{\#Params (M)} & \multicolumn{2}{c|}{\textbf{Mean Dice Score (\%)}} & \textbf{Arch.} \\
\midrule
\textbf{Dataset} & & \textbf{ACDC} & \textbf{AMOS} & \\
\midrule
UNETR \cite{hatamizadeh2022unetr} & 146.59 & 85.34 & 76.59 & ViT \\SwinUNETR \cite{hatamizadeh2021swin} & 62.19 & 91.29 & 83.81 & ViT \\
SwinUNETRV2 \cite{he2023swinunetr} & 72.76 & 92.01 & 86.24 & ViT \\
nnFormer \cite{10183842} & 37.69 & \textbf{92.40} & 81.55 & ViT \\
CoTr \cite{xie2021cotr} & 41.93 & 90.56 & 88.02 & ViT \\
\midrule
SAM3D \cite{10635844} & 1.88 & 90.41 & 82.48 & SAM\\
\midrule
UNETVL w/o KAN (\(K=384\)) & 94.96 & 90.14 & 86.52 & ViL\\
UNETVL w/ KAN (\(K=384\)) & 158.61 & \underline{91.59} & \underline{\textbf{88.57}} & ViL\\
\bottomrule
\end{tabular}
} \caption{Results on ACDC and AMOS post-challenge Task 2 datasets. All results, except for UNETR and SAM3D, are from the nnUNet revisited work. The bolded values represent the best results, while the underlined values highlight the better performance between the two configurations of UNETVL using the Chebyshev KAN layer or not.}
\label{q1}
\end{table}

\noindent\textbf{Ablation Studies.}
To further dissect the contributions of various components of the proposed UNETVL architecture, we conducted ablation studies (refer to Table \ref{q2}). The models with and without KAN, and with varying latent dimensions, were evaluated. As expected, the use of KAN consistently improves performance, and increasing the latent dimension results in better segmentation, albeit at the cost of higher parameter counts.

\noindent\textbf{Choice of Univariate Functions for KAN.} To rigorously assess the performance of various univariate functions in the KAN layer, we conducted experiments within a light-weight CNN-based U-KAN model \cite{li2024ukanmakesstrongbackbone}. Using 5-fold cross-validation on the ACDC dataset ensured the robustness of our evaluation. As shown in Table \ref{q3}, Chebyshev polynomials consistently demonstrated superior performance compared to MLP, B-spline, and Gaussian RBF. This highlights the Chebyshev function's capability to capture intricate anatomical structures, making it well-suited for complex medical image segmentation tasks.

\label{ssec:subhead}
\begin{table}[h]
\centering
\tiny
\resizebox{\columnwidth}{!}{ 
\begin{tabular}{l| c| c}
\toprule
\textbf{Model Configuration} & \textbf{\#Params (M)} & \textbf{Mean Dice Score (\%)} \\
\midrule
w/o KAN \& \(K=192\) & 74.36 & 85.88 \\
w/o KAN \& \(K=384\) & 94.96 & 87.83 \\
w/ KAN \& \(K=192\)  & 90.28 & 86.34 \\
w/ KAN \& \(K=384\) & 158.61 & 89.79 \\
\bottomrule
\end{tabular}
}
\caption{Ablation studies of different configurations and test on ACDC cross validation fold 1 after 100 epochs.}
\label{q2}
\end{table}

\begin{table}[h]
\centering
\tiny
\resizebox{\columnwidth}{!}{ 
\begin{tabular}{l|cccc}
\toprule
\textbf{Method} & \textbf{1} & \textbf{2} & \textbf{3} & \textbf{Avg.} \\
\midrule
MLP           & 83.75  & 83.22  & 90.65  & 85.88 \\
B-Spline      & 83.63  & 83.19  & 90.59  & 85.80 \\
Gaussian RBF  & 83.60  & 83.09  & 90.61  & 85.77 \\
Chebyshev     & \textbf{84.04} & \textbf{83.28} & \textbf{91.03} & \textbf{86.12} \\
\bottomrule
\end{tabular}
}
\caption{Dice score (\%) of different univariate function choices for KAN layer based on U-KAN \cite{li2024ukanmakesstrongbackbone} architecture. The best results for each column are highlighted in bold.}
\label{q3}
\end{table}

\section{Discussion and Conclusion}
\label{sec:foot}
The proposed UNETVL architecture, enhanced by Chebyshev KAN, demonstrates competitive performance in 3D medical image segmentation tasks. While the Dice score may not be the absolute highest, the model significantly mitigates over-segmentation, capturing finer anatomical details and providing cleaner segmentations. This capability is particularly beneficial in clinical applications where precise delineation of complex structures is critical for diagnostic and treatment planning. The Chebyshev KAN, compared to other univariate functions, excels in handling complex structures, offering a flexible and robust alternative for MLP on medical image segmentation tasks. However, the increased complexity of the Chebyshev KAN layer adds to the computational burden, particularly when scaling to larger datasets or higher-resolution images. Future work may explore optimizations to mitigate these costs.



\section{Compliance with ethical standards}


This research study was conducted retrospectively using human subject data made available in open access by Bernard et al. \cite{bernard2018deep} and Ji et al. \cite{NEURIPS2022_ee604e1b}. Ethical approval was not required as confirmed by the license attached with the open access data.




\bibliographystyle{IEEEbib}
\bibliography{unetvl}

\begin{thebibliography}{10}

\bibitem{hatamizadeh2022unetr}
Ali Hatamizadeh, Yucheng Tang, Vishwesh Nath, Dong Yang, Andriy Myronenko, Bennett Landman, Holger~R Roth, and Daguang Xu,
\newblock ``Unetr: Transformers for 3d medical image segmentation,''
\newblock in {\em Proceedings of the IEEE/CVF winter conference on applications of computer vision}, 2022, pp. 574--584.

\bibitem{ronneberger2015u}
Olaf Ronneberger, Philipp Fischer, and Thomas Brox,
\newblock ``U-net: Convolutional networks for biomedical image segmentation,''
\newblock in {\em Medical image computing and computer-assisted intervention--MICCAI 2015: 18th international conference, Munich, Germany, October 5-9, 2015, proceedings, part III 18}. Springer, 2015, pp. 234--241.

\bibitem{beck2024xlstm}
Maximilian Beck, Korbinian P{\"o}ppel, Markus Spanring, Andreas Auer, Oleksandra Prudnikova, Michael Kopp, G{\"u}nter Klambauer, Johannes Brandstetter, and Sepp Hochreiter,
\newblock ``xlstm: Extended long short-term memory,''
\newblock {\em arXiv preprint arXiv:2405.04517}, 2024.

\bibitem{alkin2024vision}
Benedikt Alkin, Maximilian Beck, Korbinian P{\"o}ppel, Sepp Hochreiter, and Johannes Brandstetter,
\newblock ``Vision-lstm: xlstm as generic vision backbone,''
\newblock {\em arXiv preprint arXiv:2406.04303}, 2024.

\bibitem{liu2024kan}
Ziming Liu, Yixuan Wang, Sachin Vaidya, Fabian Ruehle, James Halverson, Marin Solja{\v{c}}i{\'c}, Thomas~Y Hou, and Max Tegmark,
\newblock ``Kan: Kolmogorov-arnold networks,''
\newblock {\em arXiv preprint arXiv:2404.19756}, 2024.

\bibitem{bernard2018deep}
Olivier Bernard, Alain Lalande, Clement Zotti, Frederick Cervenansky, Xin Yang, Pheng-Ann Heng, Irem Cetin, Karim Lekadir, Oscar Camara, Miguel Angel~Gonzalez Ballester, et~al.,
\newblock ``Deep learning techniques for automatic mri cardiac multi-structures segmentation and diagnosis: is the problem solved?,''
\newblock {\em IEEE transactions on medical imaging}, vol. 37, no. 11, pp. 2514--2525, 2018.

\bibitem{NEURIPS2022_ee604e1b}
Yuanfeng Ji, Haotian Bai, Chongjian GE, Jie Yang, Ye~Zhu, Ruimao Zhang, Zhen Li, Lingyan Zhanng, Wanling Ma, Xiang Wan, and Ping Luo,
\newblock ``Amos: A large-scale abdominal multi-organ benchmark for versatile medical image segmentation,''
\newblock in {\em Advances in Neural Information Processing Systems}, 2022, vol.~35, pp. 36722--36732.

\bibitem{isensee2024nnu}
Fabian Isensee, Tassilo Wald, Constantin Ulrich, Michael Baumgartner, Saikat Roy, Klaus Maier-Hein, and Paul~F Jaeger,
\newblock ``nnu-net revisited: A call for rigorous validation in 3d medical image segmentation,''
\newblock in {\em International Conference on Medical Image Computing and Computer-Assisted Intervention}. Springer, 2024, pp. 488--498.

\bibitem{ss2024chebyshev}
Sidharth SS,
\newblock ``Chebyshev polynomial-based kolmogorov-arnold networks: An efficient architecture for nonlinear function approximation,''
\newblock {\em arXiv preprint arXiv:2405.07200}, 2024.

\bibitem{isensee2021nnu}
Fabian Isensee, Paul~F Jaeger, Simon~AA Kohl, Jens Petersen, and Klaus~H Maier-Hein,
\newblock ``nnu-net: a self-configuring method for deep learning-based biomedical image segmentation,''
\newblock {\em Nature methods}, vol. 18, no. 2, pp. 203--211, 2021.

\bibitem{hatamizadeh2021swin}
Ali Hatamizadeh, Vishwesh Nath, Yucheng Tang, Dong Yang, Holger~R Roth, and Daguang Xu,
\newblock ``Swin unetr: Swin transformers for semantic segmentation of brain tumors in mri images,''
\newblock in {\em International MICCAI brainlesion workshop}. Springer, 2021, pp. 272--284.

\bibitem{he2023swinunetr}
Yufan He, Vishwesh Nath, Dong Yang, Yucheng Tang, Andriy Myronenko, and Daguang Xu,
\newblock ``Swinunetr-v2: Stronger swin transformers with stagewise convolutions for 3d medical image segmentation,''
\newblock in {\em International Conference on Medical Image Computing and Computer-Assisted Intervention}. Springer, 2023, pp. 416--426.

\bibitem{10183842}
Hong-Yu Zhou, Jiansen Guo, Yinghao Zhang, Xiaoguang Han, Lequan Yu, Liansheng Wang, and Yizhou Yu,
\newblock ``nnformer: Volumetric medical image segmentation via a 3d transformer,''
\newblock {\em IEEE Transactions on Image Processing}, vol. 32, pp. 4036--4045, 2023.

\bibitem{xie2021cotr}
Yutong Xie, Jianpeng Zhang, Chunhua Shen, and Yong Xia,
\newblock ``Cotr: Efficiently bridging cnn and transformer for 3d medical image segmentation,''
\newblock in {\em Medical Image Computing and Computer Assisted Intervention--MICCAI 2021: 24th International Conference, Strasbourg, France, September 27--October 1, 2021, Proceedings, Part III 24}. Springer, 2021, pp. 171--180.

\bibitem{10635844}
Nhat-Tan Bui, Dinh-Hieu Hoang, Minh-Triet Tran, Gianfranco Doretto, Donald Adjeroh, Brijesh Patel, Arabinda Choudhary, and Ngan Le,
\newblock ``Sam3d: Segment anything model in volumetric medical images,''
\newblock in {\em 2024 IEEE International Symposium on Biomedical Imaging (ISBI)}, 2024, pp. 1--4.

\bibitem{li2024ukanmakesstrongbackbone}
Chenxin Li, Xinyu Liu, Wuyang Li, Cheng Wang, Hengyu Liu, Yifan Liu, Zhen Chen, and Yixuan Yuan,
\newblock ``U-kan makes strong backbone for medical image segmentation and generation,'' 2024.

\end{thebibliography}

\end{document}